\documentclass[runningheads]{llncs}
\usepackage{graphicx}
\usepackage{amsmath, amssymb,bm}
\usepackage{xcolor}
\usepackage{booktabs}

\usepackage[colorlinks=true,citecolor=blue,urlcolor=black]{hyperref}

\usepackage{comment}
\usepackage{bbm}
\usepackage{multirow}
\usepackage{threeparttable}
\usepackage{esvect}

\begin{document}

\title{Incorporating Temporal Prior from Motion Flow \\ for Instrument Segmentation in Minimally Invasive Surgery Video}

\titlerunning{Incorporating Temporal Prior for Surgical Instrument Segmentation}

\author{Yueming Jin\inst{1}, Keyun Cheng\inst{1}, Qi Dou\inst{2} \and Pheng-Ann Heng\inst{1,3} }
\institute{Dept. of Computer Science and Engineering, The Chinese University of Hong Kong
	\and Department of Computing, Imperial College London
	\and T Stone Robotics Institute, The Chinese University of Hong Kong}

\authorrunning{Y. Jin et al.}

\maketitle              

\begin{abstract}
Automatic instrument segmentation in video is an essentially fundamental yet challenging problem for robot-assisted minimally invasive surgery.
In this paper, we propose a novel framework to leverage instrument motion information, by incorporating a derived temporal prior to an attention pyramid network for accurate segmentation.
Our inferred prior can provide reliable indication of the instrument location and shape, which is propagated from the previous frame to the current frame according to inter-frame motion flow.
This prior is injected to the middle of an encoder-decoder segmentation network as an initialization of a pyramid of attention modules, to explicitly guide segmentation output from coarse to fine.
In this way, the temporal dynamics and the attention network can effectively complement and benefit each other.
As additional usage, our temporal prior enables semi-supervised learning with periodically unlabeled video frames, simply by reverse execution.
We extensively validate our method on the public 2017 MICCAI EndoVis Robotic Instrument Segmentation Challenge dataset with three different tasks.
Our method consistently exceeds the state-of-the-art results across all three tasks by a large margin.
Our semi-supervised variant also demonstrates a promising potential for reducing annotation cost in the clinical practice.

\end{abstract}

\section{Introduction}

With advancements of robot-assisted minimally invasive surgery, enhancing automatic context awareness of the surgical procedure is important for improving surgeon performance and patient safety. 
Segmentation of the surgical instrument plays a fundamental role for various further tasks including tool pose estimation, tracking and control.  
In addition, for augmented reality, referring a segmentation mask can prevent the overlay of rendered tissue from occluding instruments.
However, accurate instrument segmentation from surgical videos is very challenging, due to the complicated scene, blur from instrument motion, inevitable visual occlusion by blood or smoke, and various lighting conditions.
Recognizing the instrument in greater details, e.g., separating its different parts or specifying its sub-type, is even harder given the limited inter-class variance.

To meet these challenges, early methods use hand-crafted features from color and texture,
with machine learning models such as random forests and Gaussian mixture model~\cite{bouget2015detecting,rieke2016real}.
Later, convolutional neural network (CNN) based methods have demonstrated new state-of-the-art on instrument segmentation.
The ToolNet~\cite{garcia2017toolnet} uses a holistically-nested fully convolutional network, imposing multi-scale constraint of predictions.
Laina et al.~\cite{laina2017concurrent1} propose a multi-task CNN to concurrently regress the segmentation and localization.
Milletari et al.~\cite{milletari2018cfcm1} use residual CNN and integrate multi-scale features of a frame via LSTM.
Shvets et al.~\cite{shvets2018automatic} design a skip-connection model trained with transfer learning, winning the 2017 EndoVis Challenge~\cite{allan20192017}.
The existing works treat sequential data as static image, and perform segmentation purely using visual cues in single frame.

With the sequential nature, temporal information actually can provide valuable clues for video analysis, and has demonstrated benefit in other surgical tasks,
e.g., workflow recognition~\cite{jin2018sv,twinanda2017endonet}, instrument detection~\cite{sarikaya2017detection}, and pose estimation~\cite{allan20183}.
These methods either implicitly learn spatio-temporal features in a network (generally with LSTM), 
or straightforwardly take the optical flow map as an extra input channel to a network.
In addition, they only need to produce coarse predictions rather than pixel-level dense segmentation.
How to more interpretably utilize time cues and more explicitly incorporate it into a network, are of large importance to achieve an accurate segmentation.

We propose a novel framework integrating a prior derived from \textbf{m}otion \textbf{f}low into a \textbf{t}emporal \textbf{a}ttention \textbf{p}yramid network (named MF-TAPNet) for automatic instrument segmentation in minimally invasive surgery video.
Our method uses the inherent temporal clues from the instrument motion to boost results.
Specifically, we propagate the prediction mask of the previous frame, via optical flow in an unsupervised way, and infer a reliable prior indicating the instrument's location and shape in the current frame.
Next, we make explicit use of this temporal prior, by incorporating it at the bottleneck layer of a segmentation network as an initial attention map, and evolve a pyramid of attention modules. 
In this way, the sequential dynamics and the attention network can complement and progressively highlight discriminative features (or suppress irrelevant regions).
As an exciting additional usage, our method enables semi-supervised learning at periodically unlabeled video, simply by propagating the prior in reverse direction.
We evaluate our method on three different tasks of 2017 MICCAI EndoVis Challenge.
Our MF-TAPNet consistently outperforms the leaderboard methods at all tasks.
Our semi-supervised setting also achieves promising results only requiring labeling 50\% frames, which endorses potential value in clinical practice.

\section{Method}

\begin{figure}
	\centering
	\includegraphics[width=0.99\linewidth]{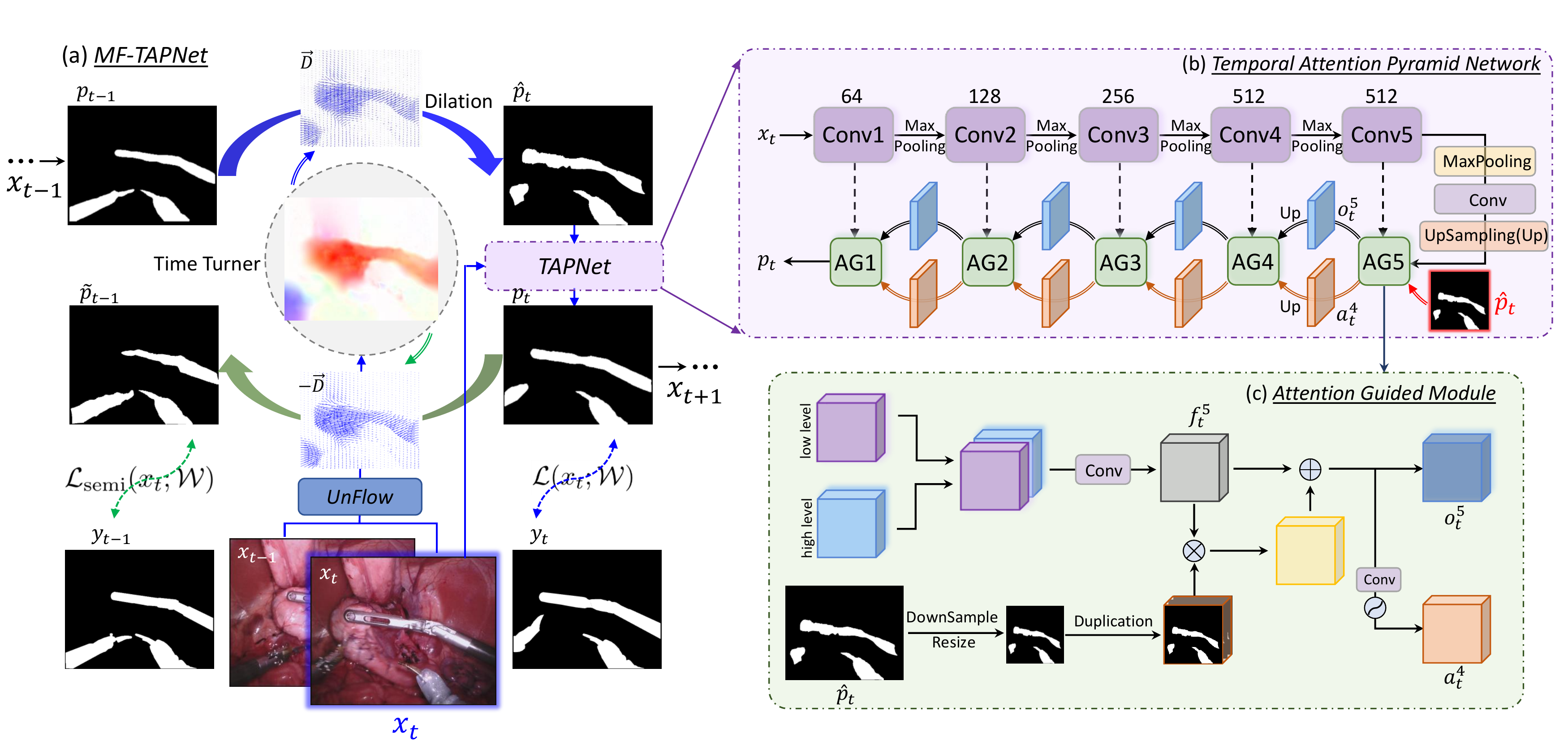}
	\caption{Illustration of the proposed (a) MF-TAPNet for surgical instrument segmentation based on motion flow, with architecture of (b) temporal attention pyramid network and (c) attention guided module presented in detail.}
	\label{fig:overview}
\end{figure}

Fig.~\ref{fig:overview} presents our proposed MF-TAPNet, which incorporates motion flow based temporal prior to a designed attention pyramid network for accurate surgical instrument segmentation from video. We elaborate each component in this section.

\subsection{Unsupervised Temporal Propagation via Motion Flow}
In surgical video, instruments performed by surgeons, usually have obvious and rich motion information.
Such valuable temporal inherence in the sequential data is unexplored in previous works on instrument segmentation.
We propose a novel temporal prior propagation strategy, named as \textbf{\textit{time turner}}, to take advantage of such domain knowledge to a large extent.

Intuitively, we argue that the motion derived from the raw images (frames in video data) also applies to their corresponding instrument masks. 
This generally shares the spirit with atlas-based segmentation, but here, our ``deformation field" is the motion flow between sequential frames in video.
In practice, we derive the apparent instrument movement using optical flow which is de-facto for motion analysis.
More specifically, we use Unflow~\cite{meister2018unflow}, a recent state-of-the-art method, to obtain a map $\vv{\bm{D}}$ of displacement vector between adjacent frame pair of $(x_{t-1}, x_t)$, showing the motion magnitude and orientation at each pixel.
In the map $\vv{\bm{D}}$, each displacement vector $\vv{\bm{d}} \! = \! [d_a,d_b]$ directs a position from frame $x_{t-1}$ to $x_t$. 
In our intuition, their instrument masks also follow the same location shift with such motion. 
Given the mask prediction $p_{t-1}$ (output from a network given input of $x_{t-1}$), we can propagate it with $\vv{\bm{D}}$ to infer a prediction for $x_t$. 
Formally, with denoting $\bm{u}_{t-1}\! = \! [u_a,u_b]$ as the position of one value in $p_{t-1}$, we infer its position in next frame as $\bm{u}_t \! = \! \bm{u}_{t-1} \! + \! \vv{\bm{d}} \! = \! [u_a\! + \! d_a,u_b \! + \!d_b]$.
With operation for all positions, we obtain the inferred mask prediction for $x_t$, which is the referred concept of \textit{temporal prior} in this paper.
We further enhance it using morphological dilation to relieve the effects of camera zoom,
The finally obtained temporal prior is denoted by $\hat{p}_t$, which is very informative and of high-quality regarding the location and shape of instrument in frame $x_t$.
Note that in multi-class segmentation, we sum the probabilities of all positive classes and get $p_{t-1}$ as a 2D map indicating non-background probability, therefore $\hat{p}_t$ is also a 2D map accordingly.
Our prior can be obtained via propagating the prediction map by computing optical flow, no matter the instrument motion is large or mild compared with background motion.
Therefore, it can be well generalizable to some unusual yet extreme conditions in surgical video, such as video clips with the large camera motion, still instruments and no instrument.

\subsection{Temporal Prior Driven Attention Pyramid Network}

Way of incorporating the temporal prior $\hat{p}_t$ provided by the \textit{time turner} is crucial for taking great advantage of it.
In this regard, we design a \textbf{t}emporal \textbf{a}ttention \textbf{p}yramid \textbf{net}work (TAPNet) which consists of multi-stage attention guided (AG) modules.
It injects the prior at the encoder-decoder bottleneck and progressively learns attention guide-map pyramid in coarse-to-fine, see Fig.~\ref{fig:overview} (b). The temporal prior serves as initialization of the series of attentions, and forms the essential focus throughout the pyramid.
Some previous methods may also use multi-stage attention, however, most works implicitly learn attention maps from home-grown features within a network~\cite{Chen2018multiview1,oktay2018attention}.
Our TAPNet is explicitly driven by the distinct temporal prior,
making the model precisely focus on the instrument regions and hence the benefit of attention pyramid is maximized. 

We first elaborate the operation inside an AG module in Fig.~\ref{fig:overview} (c), with example of the most coarse one (AG5) where prior $\hat{p}_t$ is incorporated.
In the segmentation task, we use skip connection to concatenate low/high-level features, followed by $1 \! \times \!1$ convolution producing $f_t^5$.
We first downsample $\hat{p}_t$, and then duplicate it to the same channels as $f_t^5$.
Next, we conduct element-wise multiplication between $f_t^5$ and the processed $\hat{p}_t$, to extract features from those spatial locations recognized in temporal prior.
The result is resummed with $f_t^5$, outputting a representation with enlarged instrument-related activation and necessary visual context.
It is forwarded to a $3 \! \times \! 3$ convolution and a Sigmoid function to generate the attention map for next stage AG.
Formally, for the $i$-th AG, output $o_t^i$ and attention map $a_t^{i-1}$ for its following module are obtained with:
\begin{equation}
\centering
\begin{gathered}
o_t^i = f_t^i + a_t^i \odot f_t^i, ~~~
a_t^{i-1} = \textit{Sigmoid} ~ ( \text{Conv}  (o_t^i  ; \omega)).
\end{gathered}
\end{equation}
Both $o^i_t, a_t^{i-1}$ are upsampled by interpolation before forwarding to the next stage.
Overall, we stack 5 AG modules in pyramid to gradually decode coarse features guided by attention maps, and finally obtain the dense prediction $p_t$ for frame $x_t$.
With denoting the label mask of frame $x_t$ by $y_t$, we adopt weighted cross-entropy loss for multi-class segmentation, computing from all pixels in frame $x_t$:
\begin{equation}
\mathcal{L}(x_t ; \mathcal{W}) =  \sum -  \alpha \cdot \log \mathcal{P}(y_t|x_t, \hat{p}_{t} ), ~ \text{where} ~ \hat{p}_t \! = \! \vv{\bm D}(p_{t-1}|x_{t-1}, x_{t} ).
\label{eq:full}
\end{equation}
Similarly, prediction $p_t$ of frame $x_t$ is also used to infer $\hat{p}_{t+1}$ for its future frame $x_{t+1}$.
To the end, a beneficial circulation for the entire network training is formed, to sequentially produce accurate segmentation masks of the entire surgical video. For the very beginning frame $x_0$, its prior is set as zero, but this rare case cannot affect learning. The optical flow at \textit{time turner} is precomputed, so with $p_{t-1}$ from the network, we can compute $\hat{p}_t$ in real-time during training.

\subsection{Semi-supervision via Reverse Time Turner}

Annotating medical data is time-consuming and laborious, especially for surgical video with high frequency.
Excitingly, our method enables semi-supervised learning with fewer annotations using \textit{time turner}.
This is achieved by leveraging the sequential consistency to transfer the prediction of unlabeled
frame to that of the adjacent frame whose label is available for loss calculation.

With a video having $T$ frames as $\bm{x} \! \! = \! \! \{ x_{0}, x_{1}, \ldots, x_{T-1} \}$, we assume that $\bm{x}$ is labeled with intervals, e.g., only $\{ x_{0}, x_{2}, x_4, \ldots \}$ being labeled.
This is a reasonable setting in clinical practice because it is easier for surgeons to perform low hertz labeling.
The whole data therefore consists of labeled subset $\mathcal{V} \! = \! \{x_k \}_{k=2n}$ and unlabeled subset $\mathcal{U} \! = \! \{x_k \}_{k=2n+1}$.
If frame $x_t$ is unlabeled,
we simply execute our \textit{time turner} in a reverse direction, to transfer its prediction $p_t$ into $\tilde{p}_{t-1}$ which is corresponding to frame $x_{t-1}$.
Note that a reverse flow $-\vv{\bm{\mathcal{D}}}$ is easily obtained with element-wise negative of $\vv{\bm{\mathcal{D}}}$, without extra computation, see green arrow in Fig.~\ref{fig:overview} (a).
The $\tilde{p}_{t-1}$ is expected to well overlap with label of $x_{t-1}$, given inherent motion consistency.
Hence, we can borrow $y_{t-1}$ to calculate semi-supervised cross entropy for $x_t$, as:
\begin{equation}
\mathcal{L}_{\text{semi}}(x_t ; \mathcal{W}) =  \sum - \beta \cdot ( y_{t-1} \cdot  \log \tilde{p}_{t-1}), ~\text{where} ~ \tilde{p}_{t-1} \! = \! -\vv{\bm D} (p_t|x_{t-1},x_t).
\label{eq:semi}
\end{equation}
Overall, the training uses the supervised loss in Eq.(\ref{eq:full}) if a frame $x_t$ is labeled, otherwise it uses the semi-supervised loss in Eq.(\ref{eq:semi}).
By encouraging the temporal consistent predictions, the bi-directional use of \textit{time turner} effectively benefits the network learning.
Our semi-supervision enabled by motion flow is inherently general and can be applicable for other medical video analysis tasks.

\section{Experiments}

\textbf{Dataset and Evaluation Metrics.}
We validate the proposed framework on the public dataset of Robotic Instrument Segmentation from the 2017 MICCAI EndoVis  Challenge~\cite{allan20192017}.
It consists of $10$ video sequences of abdominal porcine procedures.
Each video contains 300 frames obtained at sampling frequency of 2\textit{Hz} and a high resolution of $1280 \! \times \! 1024$.
Specifically, $8 \times 225$-frame videos are used for training, while the remaining $8 \times 75$-frame videos and another $2 \times 300$-frame videos are used for testing; the ground-truth of test data is held-out by challenge organizer.
There are three sub-tasks, i.e. binary instrument (2 classes), instrument part (4 classes), instrument type (8 classes), gradually fine-grained segmentation of an instrument.
The challenge report~\cite{allan20192017} describes more details of the difficulties in the tasks.
For direct and fair comparison, we follow the same evaluation manner as TernausNet~\cite{shvets2018automatic} (challenge winner),
\begin{table}
	\centering
	\caption{Comparison of instrument segmentation results on three tasks (mean$\pm$std).}
	\resizebox{1\textwidth}{!}{
		\begin{tabular}{c|cc|cc|cc}
			\toprule
			\multirow{2}{*}{Methods}    & \multicolumn{2}{c}{Task1: Binary segmentation}  & \multicolumn{2}{|c}{Task2: Part segmentation}  & \multicolumn{2}{|c}{Task3: Type segmentation} \\
			\cmidrule{2-3} \cmidrule{4-5}  \cmidrule{6-7}
			& IoU (\%)       & Dice (\%) ~ & IoU (\%)       & Dice (\%)  ~ & IoU (\%)      & Dice (\%)  \\ \hline
			U-Net~\cite{ronneberger2015u1}             & ~ $75.44\pm{18.18}$ ~  &  $84.37\pm{14.58}$ ~ & ~ $48.41\pm{17.59}$ ~ & $60.75\pm{18.21}$ ~ & ~ $15.80\pm{15.06}$ ~  & $23.59\pm{19.87}$  \\ 
			TernausNet~\cite{shvets2018automatic}        & $83.60\pm{15.83}$  & $90.01\pm{12.50}$ ~ & $65.50\pm{17.22}$ & $75.97\pm{16.21}$ ~ & $33.78\pm{19.16}$ & $44.95\pm{22.89}$ \\ 
			U-NetPlus~\cite{hasan2019u}         & $83.75\pm{15.36}$  & $90.19\pm{11.77}$ ~ & $65.75\pm{16.74}$ & $76.25\pm{15.54}$ ~ & $34.19\pm{15.06}$ & $45.32\pm{19.86}$ \\ \hline
			PlainNet             & ~ $81.86\pm{15.85}$ ~  &  $88.96\pm{12.98}$ ~ & ~ $64.73\pm{17.39}$ ~ & $73.53\pm{16.98}$ ~ & ~ $34.57\pm{21.93}$ ~  & $44.64\pm{25.16}$  \\ 
			TAPNet             & $84.01\pm{16.93}$  & $90.46\pm{13.56}$ ~ & $65.84\pm{16.91}$ & $76.12\pm{16.75}$ ~ & $34.23\pm{19.63}$ & $45.50\pm{22.55}$ \\ 
			$\pmb{\text{MF-TAPNet (Ours)}}$      & $\pmb{87.56\pm{16.24}}$  & $\pmb{93.37\pm{12.93}}$ ~ & $\pmb{67.92\pm{16.50}}$ & $\pmb{77.05\pm{16.17}}$ ~ & $\pmb{36.62\pm{22.78}}$ & $\pmb{48.01\pm{25.64}}$ \\  \hline \hline
			MF-TAPNet (50\%)            & $79.31\pm{17.13}$  & $87.18\pm{13.68}$ ~ & $56.01\pm{15.59}$ & $68.13\pm{15.44}$ ~ & $28.47\pm{23.41}$ & $38.39\pm{25.88}$ \\ 
			$\pmb{\text{Semi-MF-TAPNet}}$ (50\%)      & $\pmb{80.03\pm{16.87}}$  & $\pmb{88.07\pm{13.15}}$ ~ & $\pmb{56.72\pm{16.12}}$ & $\pmb{68.51\pm{16.11}}$ ~ & $\pmb{30.04\pm{19.79}}$ & $\pmb{41.01\pm{23.81}}$ \\ 
			\bottomrule
		\end{tabular}
	}
	\label{tab:quant}
\end{table}
by using 4-fold cross-validation with the same splits of $8 \times 225$ released training data.
We also use the same evaluation metrics as~\cite{shvets2018automatic}, i.e., 
1) mean intersection-over-union (IoU), which is also used in MICCAI EndoVis Challenge to evaluate participants,
and 2) Dice coefficient (Dice), which is another common metric for segmentation.
\\
\\
\textbf{Implementation Details.}
We reduce the resolution to $640 \! \times \! 512$ to save memory.
We train models using an Adam optimizer~\cite{kingma2014adam}, with learning rates initialized as $3e\!-\!5$, $3e\!-\!5$ and $2e\!-\!5$ respectively for binary, part and type segmentation tasks.
Our framework is implemented in PyTorch with 4 NVIDIA Titan Xp GPUs for training.
The multiple GPUs enable the network to be trained at batch size of $8$.
The backbone of our network is VGG11~\cite{simonyan2014very} with 5 scales of downsampling, and deeper networks did not yield much better results in experiments, so we stick to VGG11 for the sake of real-time efficiency during surgery. The code is available at \url{https://github.com/keyuncheng/MF-TAPNet}.
\\
\\
\textbf{Comparison with State-of-the-art Methods.}
We first compare our method with the state-of-the-art results in challenge on three tasks.
Table~\ref{tab:quant} lists the performance of U-Net~\cite{ronneberger2015u1} (results quoted from~\cite{shvets2018automatic}),
TernausNet~\cite{shvets2018automatic}, 
and latest reported U-NetPlus~\cite{hasan2019u} (an enhanced U-Net with batch normalized encoders and nearest neighbor interpolation).
We see that MF-TAPNet consistently outperforms all other methods across all three tasks.
Our IoU exceeds the challenge winner by $3.96\%$ at binary segmentation, $2.42\%$ at part segmentation, and $2.84\%$ at type segmentation.
Though \cite{shvets2018automatic} and \cite{hasan2019u} develop advanced strategies to enhance a network,
our method is superior by using temporal prior to explicitly provide a reliable guidance, which helps the network learn to focus on regions of interest.
The improvement is more obvious for binary segmentation, 
because our prior also aggregates probabilities from all positive classes.
Under such homologous guidance, we achieve the highest Dice score $93.37\%$, which is useful for context-aware robot-assisted surgery.
Meanwhile, there still exists a big room to boost type segmentation performance, even though we set the highest among existing methods.
\begin{figure}
	\centering
	\includegraphics[width=0.8\linewidth]{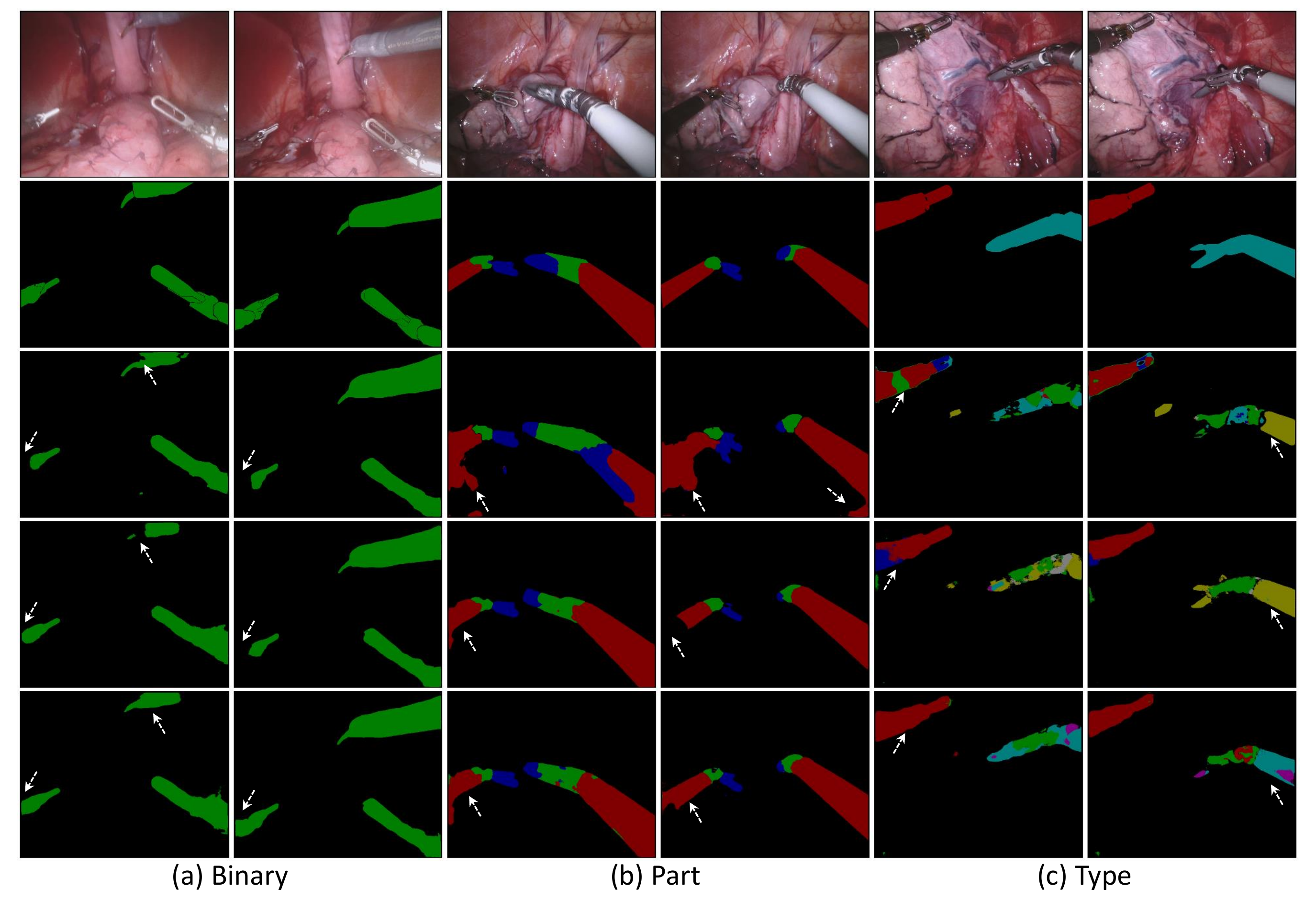}
	\caption{Typical results for instrument (a) binary segmentation (instrument and background tissues), (b) part segmentation (shaft, wrist and jaws), (c) type segmentation (different yet looking quite similar instruments).
		From top to bottom, for each task, we present two continuous video frames and their corresponding ground truth, with segmentation results using PlainNet, TAPNet and our proposed MF-TAPNet.
	}
	\label{fig:show}
\end{figure}
This reflects the natural great challenge in this task, i.e. the extremely similar appearance (shape and intensity value) in different fine-grained types.
Our method is extensible for further improvement by inferring a multi-class prior, while such extension is limited for other methods.
Last, as our method relies on the temporal consistency, 
its efficacy may degrade when unexpected motion appears, resulting in slightly higher standard deviations.
This can be alleviated as advancements of more stable surgical robots.
\\
\\
\textbf{Effectiveness of Temporal Prior and Motion Flow.}
We investigate effectiveness of key components in our MF-TAPNet.
Table~\ref{tab:quant} also lists the results of three ablation settings:
1) a plain encoder-decoder as baseline (PlainNet);
2) our TAPNet, but directly use the previous frame's prediction $p_{t-1}$ as temporal prior;
3) our entire framework at fully-supervised learning.
The network backbone is unchanged for different settings for clear comparison.
We observe that TAPNet performs better than PlainNet, especially for binary segmentation (1.50\% higher Dice) and part segmentation (2.59\% higher Dice).
This shows that, explicitly incorporating a temporal prior can provide powerful guidance, even with the rough prediction from previous frame.
Accordingly, our TAPNet can pyramidally refine the guidance and gradually concentrate on segmenting attentive objects. 
Here, our TAPNet can already achieve comparable results with the state-of-the-art methods.
More importantly, our MF-TAPNet further largely increases performances for all tasks (averagely 2.67\% IoU).
It demonstrates that after involving motion dynamics derived in \textit{time turner}, the prior presents much higher quality and can convey more accurate shape and location in current frame.
Some visual results are shown in Fig.~\ref{fig:show}.
MF-TAPNet can achieve complete and consistent segmentations, and largely suppress the irrelevant and incorrect regions.
\\
\\
\textbf{Semi-supervised Variant Enabled by Time Turner.}
We conduct experiment with the variant of semi-supervised learning.
Our setting is that the data are labeled at an interval of 2, resulting in $50\%$ frames having labels.  
In Table~\ref{tab:quant}, 
we see that our semi-supervised loss (i.e., approximating the prediction of an unlabeled frame towards a reasonable label) can better confront the performance drop at sparse annotation, compared with an ordinary training of MF-TAPNet with $50\%$ labeled data.
This is a bonus from our \textit{time turner} with interpretable meanings, and a simply reverse execution invokes a promising potential to reduce annotation cost which is very valuable in clinical surgery.

\section{Conclusion}

We propose a novel framework to incorporate temporal information pyramidally in a network for automatic instrument segmentation from robot-assisted surgery.
Our method consistently outperforms the state-of-the-art methods across all the three tasks on the 2017 MICCAI EndoVis Challenge dataset, by a large margin.
Our temporal prior enables semi-supervised learning simply by reverse execution.
The achieved outstanding results, and demonstrated potentials for extension and label efficiency, endorse a promising value of our method in clinical intervention.
\\
\\
\textbf{Acknowledgments.} The work was partially supported by HK RGC TRS project T42-409/18-R, HK RGC project CUHK14225616, and CUHK T Stone Robotics Institute, CUHK. Yueming Jin is funded by the HK Ph.D. Fellowship.

%
%
%
%

\bibliographystyle{splncs04}
\bibliography{refs}

\end{document}